\documentclass[runningheads]{llncs}

\usepackage{graphicx}
\usepackage{amsmath,amssymb} 
\usepackage{color}

\usepackage[utf8]{inputenc}        
\usepackage[T1]{fontenc}    
\usepackage{hyperref}       
\usepackage{url}            
\usepackage{booktabs}       
\usepackage{amsfonts}       
\usepackage{nicefrac}       
\usepackage{microtype}      
\usepackage{cleveref}
\usepackage{pdfpages}
\usepackage{float}
\usepackage{hhline}
\usepackage{array}
\usepackage{varwidth}
\usepackage{multirow}
\restylefloat{table}
\usepackage{makecell}
\usepackage{tabularx}
\usepackage{amsmath}
\usepackage[linesnumbered,ruled]{algorithm2e}
\usepackage{subfig}
\usepackage{graphicx}

\newcolumntype{Y}{>{\centering\arraybackslash}X}
\newcolumntype{s}{>{\hsize=.2\hsize}Y}
\newcolumntype{t}{>{\hsize=.5\hsize}Y}
\newcolumntype{u}{>{\hsize=.5\hsize}X}
\newcolumntype{b}{X}

\newcommand{\newtext}[1]{\textcolor[rgb]{0,0,0}{#1}}  
\newcommand{\flagtext}[1]{\textcolor[rgb]{0,0,0}{#1}}   
\newcommand{\deltatext}[1]{\textcolor[rgb]{0.0,0,0}{#1}} 

\begin{document}

\def\ECCV18SubNumber{2698}  

\title{Out-of-Distribution Detection Using an Ensemble of Self Supervised Leave-out Classifiers} 

\titlerunning{OOD detection using an ensemble of leave-out classifiers}

\author{Apoorv Vyas\inst{1}\inst{3}\thanks{Equal contribution. Work done when the authors were working at Intel labs.} \and Nataraj Jammalamadaka\inst{1}\inst{\star} \and Xia Zhu\inst{2}\inst{\star}\and Dipankar Das\inst{1} \and Bharat Kaul\inst{1} \and Theodore L. Willke\inst{2}
}
\authorrunning{Apoorv Vyas et al}

\institute{
Intel labs, Bangalore, India\\
\email{apoorv.vyas}@idiap.ch
\email{natraj.j}@gmail.com 
\email{\{dipankar.das,bharat.kaul\}}@intel.com\\
\and
Intel labs, Hillsboro, OR 97124, USA\\
\email{\{xia.zhu,ted.willke\}}@intel.com\\
\and 
Idiap Research Institute, Switzerland\\
 }

\maketitle              

\begin{abstract} 
As deep learning methods form a critical part in commercially important applications such as autonomous driving and medical diagnostics, it is important to reliably detect out-of-distribution (OOD) inputs while employing these algorithms. In this work, we propose an OOD detection algorithm which comprises of an ensemble of classifiers. We train each classifier in a self-supervised manner by leaving out a random subset of training data as OOD data and the rest as in-distribution (ID) data. We propose a novel margin-based loss over the softmax output which seeks to maintain at least a margin $m$ between the average entropy of the OOD and in-distribution samples. In conjunction with the standard cross-entropy loss, we minimize the novel loss to train an ensemble of classifiers. We also propose a novel method to combine the outputs of the ensemble of classifiers to obtain OOD detection score and class prediction. Overall, our method convincingly outperforms Hendrycks \emph{et al.}~\cite{hendrycks2017baseline} and the current state-of-the-art ODIN~\cite{liang2018odin} on several OOD detection benchmarks. 

\end{abstract}

\section{Introduction}
\label{sec:introduction}
Deep learning has significantly improved the performance of machine learning systems in fields such as computer vision, natural language processing, and speech. In turn, these algorithms are integral in commercial applications such as autonomous driving, medical diagnosis, and web search. In these applications, it is critical to detect sensor failures, unusual environments, novel biological phenomena, and cyber attacks. To accomplish this, systems must be capable of detecting when inputs are anomalous or out-of-distribution (OOD). In this work, we propose an out-of-distribution detection method for deep neural networks and demonstrate its performance across several out-of-distribution classification tasks on the state-of-the-art deep neural networks such as DenseNet\cite{huang2016densenet} and Wide ResNet(WRN)\cite{zagoruyko2016wrn}.

%
%
We propose a novel margin-based loss term, added to cross-entropy loss over in-distribution samples, which maintains a margin of at least $m$ between the average entropy of OOD and ID samples respectively. We propose an ensemble of $K$ leave-out classifiers for OOD detection. The training dataset with $N$ classes is partitioned into $K$ subsets such that the classes of each partition are mutually exclusive with respect to each other. Each classifier samples one of the $K$ subsets without replacement as \textit{out-of-distribution training data} and the rest of the $K-1$ subsets as \textit{in-distribution training data}. We also propose a new OOD detection score which combines both softmax prediction score and entropy with temperature scaling~\cite{liang2018odin}. We demonstrate the efficacy of our method on standard benchmarks proposed in ODIN~\cite{liang2018odin}  and \textit{outperform them}. Our contributions are (i) proposing a novel loss for OOD detection, (ii) \flagtext{demonstrating a self-supervised OOD detection method}, and (iii) moving the state-of-the-art by outperforming the current best methods.
 
The rest of the paper is organized as follows. Section~\ref{sec:relatework} describes the previous work on the OOD detection. Section~\ref{sec:method} describes our method in detail. Section~\ref{sec:experiments} describes various evaluation metrics to measure the performance of OOD detection algorithms. The ablation results of various design choices and hyper-parameters are also presented. We then compare our method against the recently proposed ODIN algorithm~\cite{liang2018odin} and demonstrate that it outperforms it on various OOD detection benchmarks. Finally, section~\ref{sec:conclusion} discusses observations about our method, future directions and conclusions.

\section{Related Work}
\label{sec:relatework}
 
Traditionally, based on the availability of the data labels, OOD detection methods can be categorized into supervised~\cite{supervisedAnomaly}, semi-supervised~\cite{semisupervisedAnomaly} and unsupervised methods~\cite{unsupervisedAnomaly},~\cite{survey09}. \newtext{All these classes of methods have access to the OOD data while training but differ in access to labels.} It is assumed that the classifier has labels for normal as well as OOD classes during training for supervised OOD detection, while labels for only the normal classes are available in case of semi-supervised methods, and no labels are provided for unsupervised OOD detection methods which typically rely on the fact that anomalies occur in much less frequency than normal data. \flagtext{ Our method is able to detect anomalies in test OOD datasets the very first time it encounters them during testing. We use one OOD dataset as validation set to search for hyper-parameters.}


Notable OOD detection algorithms which work in the same setting as ours are isolation forests~\cite{iForest}, Hendrycks and Gimpel~\cite{hendrycks2017baseline}, ODIN~\cite{liang2018odin} and Lee \emph{et.al,}~\cite{lee2018training}. 
The work reported in isolation forests~\cite{iForest} exploits the fact that anomalies are scarce and different and while constructing the isolation tree, it is observed that the anomalous samples appear close to the root of the tree. These anomalies are then identified by measuring the length of the path from the root to a terminating node; the closer a node is to the root, the higher is its chance of representing an OOD. Hendrycks and Gimpel~\cite{hendrycks2017baseline} is based on the observation that prediction probability of incorrect and out-of-distribution samples tends to be lower than the prediction probability of correct samples. 
Lee \emph{et al.} modify the formulation of generative adversarial networks~\cite{NIPS2014_5423} to generate OOD samples for the given in-distribution. They achieve this by simultaneously training GAN~\cite{NIPS2014_5423} and standard supervised neural network. The joint loss consists of individual losses and an additional connecting term which reduces the KL divergence between the generated sample's softmax distribution and the uniform distribution.

\newtext{Another set of related works are open set classification methods{~\cite{RuddJSB18},~\cite{ScheirerJB14},~\cite{ScheirerRSB13}, ~\cite{BendaleB15},~\cite{BendaleB16}}. Scheirer {\it et.al,}~\cite{ScheirerRSB13} introduces and formalizes ``open space risk'' which intuitively is the risk associated with labeling those areas in the output feature space as positive where there is no density support from the training data. Thus the approximation to the ideal risk is defined as a linear combination of ``open space risk'' and the standard ``empirical risk''. Bendale and Boult~\cite{BendaleB15} extend the definition of ``open set risk'' to open world recognition where the unknown samples are not static set. The open world recognition defines a {\it multi-class open set recognition function}, a {\it labeling process} and an {\it incremental learning function}. The {\it multi-class open set recognition function} detects novel classes which are labeled using the {\it labeling process} and finally are fed to {\it incremental learning function} which updates the model. The OSDN~\cite{BendaleB16} work proposes openMax function which extends the softmax function by adding an additional unknown class to the classification layer. The value for unknown class is computed by taking the weighted average of all other classes. 
The weights are obtained from Weibull distribution learnt over the pairwise distances between penultimate activation vectors (AV) of the top farthest correctly classified samples. 
For an OOD test sample these weights will be high while for an in-distribution sample these scores will be low. The final activation vector is re-normalized using softmax function. 
}

The current state-of-the-art is ODIN~\cite{liang2018odin} which proposes to increase the difference between the maximum softmax scores of in-distribution and OOD samples by (i) calibrating the softmax scores by scaling the logits that feed into softmax by a large constant (referred to as temperature) and (ii) pre-processing the input by perturbing it with the loss gradient. \newtext{ODIN~\cite{liang2018odin}
demonstrated that at high temperature values, the softmax score for the predicted class is proportional to the relative difference between largest unnormalized output (logit) and the remaining outputs (logits). Moreover, they empirically showed that the difference between the largest logit and the remaining logits is higher for the in-distribution images than for the out-of-distribution images. Thus temperature scaling pushes the softmax scores of in- and out-of-distribution images further apart when compared to plain softmax. Perturbing the input image through gradient ascent w.r.t to the score of predicted label was demonstrated~\cite{liang2018odin} to have stronger effect on the in- distribution images than that on out-of-distribution images, thereby, further pushing apart the softmax scores of in- and out-of-distribution images. We leverage the effectiveness of both these methods.} The proposed method outperforms all the above methods by considerable margins.

\section{Out-of-Distribution (OOD) Classifier}
\label{sec:method}

In this section, we introduce three important components of our method: entropy based margin-loss function (\ref{sec:loss}), training ensemble of leave-out classifiers (\ref{sec:leave-out}), and OOD detection scores (\ref{sec:ood-detector}).

    
        
        
        

\begin{algorithm}[th]

    \SetKwInOut{Input}{Input}
    \SetKwInOut{Output}{Output}

    \Input{Training Data $X$, Number of classes $N$,
    $K$ Partitions, \deltatext{$\delta$ accuracy bound, Validation OOD Data $X_{valOOD}$ }}
    \Output{$K$ Leave Out Classifiers}
    
    
    \For{$i \gets 1$ \textbf{to} $K$} {
        $X_{ood} \gets X_i $, $X_{in} \gets X - X_i $\;
        
        \While{ Not Converged }{
            $ood\_batch \gets $  Sample\ OOD\ minibatch\;
            $in\_batch \gets $ Sample\ in-distribution minibatch\;
            update the classifier $F_i$ by minimizing loss (Equation~\ref{loss}) using SGD\;
            \deltatext{save model with least OOD error on $X_{valOOD}$ within $\delta$ accuracy of current best accuracy.};
        }

    }
    \Return{$\{F_i\}$}\;
    \caption{Algorithm to train $K$ Leave Out Classifiers}
    \label{algo2}
\end{algorithm}

\subsection{Entropy based Margin-Loss}
\label{sec:loss}
Given a labeled set $(x_i \in X_{in},y_i \in Y_{in})$ of in-distribution (ID) samples and $(x_o \in X_{ood})$ of out-of-distribution (OOD) samples, we propose a novel loss term in addition to the standard cross-entropy loss on ID samples. This loss term seeks to maintain a margin of at least $m$ between the average entropy of OOD and ID samples. Formally, a multi-layer neural network $F:x \rightarrow p$ which maps an input $x$ to probability over classes and parametrized by $W$ is learned by minimizing the margin-loss over the difference of average entropies over OOD samples and ID samples, and cross entropy loss on ID samples. The loss function is given by Equation~\ref{loss}, 
\begin{equation}
\resizebox{.9\hsize}{!}{$\mathcal{L}=-\frac{1}{|X_{in}|}\sum\limits_{x_i \in X_{in}}\text{log}(F_{y_i}(x_i)) + \beta*\text{max}\left( m + \frac{\sum\limits_{x_i \in X_{in}} H(F(x_i))}{\vert X_{in}\vert}-\frac{\sum\limits_{x_o \in X_{ood}} H(F(x_o))}{\vert X_{ood}\vert } ,0\right)\label{loss}$}
\end{equation}
where $F_{y_i}(x_i)$ is the predicted probability of sample $x_i$ whose ground truth class $y_i$, $H(\cdot)$ is the entropy over the softmax distribution, $m$ is the margin and $\beta$ is the weight on margin entropy loss. 

The new loss term evaluates to its minimum value \emph{zero} when the difference between the average entropy of OOD and ID samples is greater than the margin $m$. For ID samples, the entropy loss encourages the softmax probabilities of non ground-truth classes to decrease and the cross-entropy loss encourages the softmax probability of ground-truth class to increase. For OOD samples, the entropy loss encourages the probabilities of all the classes to be equal. When the OOD entropy is higher than ID entropy by a margin $m$, the new loss term evaluates to \emph{zero}. Our experiments suggest that maximizing OOD entropy leads to overfitting. Bounding the difference of average entropies of ID samples and OOD samples entropies with margin has helped in preventing overfitting, and thus is better for model generalization\cite{lecun2006energy}.  



\begin{algorithm}

    \SetKwInOut{Input}{Input}
    \SetKwInOut{Output}{Output}

    \Input{Test Image $x_t$, $K$ leave-out Classifiers $F_i, i \in {1,...,K}$ and their temperature scaled versions $F_i(x_t;T)$, perturbation factor $\epsilon$, number of classes $N$}
    \Output{Class prediction $C_t$, OOD score $O_t$}
    
    $S_t \gets \{0\}^N$, $O_t \gets 0$\;
    \For{$i \gets 1$ \textbf{to} $K$} {
        $S_t \gets S_t + F_i(x_t) $\;\
        $\hat{x}_t \gets x_t - \epsilon * \text{sign}(\frac{\partial H(F_i(\hat{x}_t;T))}{\partial x_t})  $\;\
        $O_t \gets O_t + \text{max}_N(F_i(\hat{x}_t;T)) - H(F_i(\hat{x}_t;T))$\;
    }
    $C_t \gets \text{argmax}(S_t)$\;
    \Return{$C_t$, $O_t$}\;
    \caption{Algorithm for OOD Detection using $K$ Leave Out Classifiers}
    \label{algo3}
\end{algorithm}

\subsection{Training Ensemble of leave-out classifiers}
\label{sec:leave-out}
Given an in-distribution training data \newtext{$X$} which consists of $N$ classes, the data is divided into $K$ partitions \newtext{$X_i,i\in \{1,...,K\}$} such that the classes of each partition are mutually exclusive to all other partitions. A set of $K$ classifiers are learned where classifier $F_i, i \in \{1,...,K\}$ uses the partition \newtext{$X_i$} as OOD data $X_{ood}$ and rest of the data \newtext{$X-X_i$} as in-distribution data $X_{in}$. A particularly simple way of partitioning the classes is to divide them into partitions with equal number of classes. For example, dividing a dataset of $N = 100$ classes into $K = 5$ random and equal partitions gives us a partition with size of $20$ classes. Each of the $K$ classifiers would then use \newtext{$20$ classes for OOD and $80$ classes as ID.} Each classifier $F_i$ is learned by minimizing the proposed \newtext{margin entropy} loss (eqn~\ref{loss}) using the assigned OOD and ID data. \deltatext{During the training, we also assume a small number of out-of-distribution images to be available as a validation dataset. At every epoch, we save the model with best  OOD detection rate on this small OOD validation data and within a $\delta$ accuracy bound of the current best accuracy.} The complete algorithm for training the leave-out classifiers in presented in Algorithm~\ref{algo2}.



\subsection{OOD Detection Score for Test Image}
\label{sec:ood-detector}
At the time of testing, an input image is forward propagated through all the $K$ leave-out classifiers and the softmax vectors of all the networks are remapped to their original class indices. For the left-out classes, a score of zero is assigned. For classification of an input sample, first the softmax vectors of all the classifiers are averaged and the class with the highest averaged \newtext{softmax} score is considered as the prediction. For the OOD detection score, for each of the $K$ classifiers, we first compute \newtext{both} the maximum value and negative entropy of the softmax vectors \newtext{with temperature scaling}. We then compute the average of all these values to obtain the OOD detection score.

An in-distribution sample with class labels \newtext{$y_i$} acts as an OOD for exactly one of the $K$ classifiers. This is because the classes are divided into $K$ mutually exclusive partitions and class \newtext{$y_i$} can be part of only one of these partitions. When an in-distribution sample $(x_i,y_i)$ is forward propagated through these $K$ classifiers, we would expect the negative entropy and maximum softmax score to be high for $K-1$ classifiers where it was sampled as in-distribution dataset. However, for an OOD sample $x_o$ we expect the negative entropy and maximum softmax score to be relatively low for all the $K$ classifiers. We thus expect a higher OOD detection score for ID samples than the OOD samples thus differentiating them. 

Following the work of ODIN~\cite{liang2018odin}, we use both temperature scaling and input preprocessing while testing. In temperature scaling, the logits feeding into softmax layer are scaled by a constant factor $T$. It has been established that temperature scaling can calibrate the classification score and in the context of OOD detection~\cite{liang2018odin}, \newtext{it pushes the softmax scores of in- and out-of-distribution samples further apart when compared to plain softmax.} We modify input preprocessing by perturbing over entropy loss instead of cross-entropy loss used by ODIN~\cite{liang2018odin}. Perturbing using the entropy loss decreases the entropy of the ID samples much more than the OOD samples. For an input test image $x_t$, after it is forward propagated through the neural network $F_i$, the gradient of the entropy loss with respect to $x_t$ is computed and the input is perturbed with Equation~\ref{preprocessing}. The OOD detection score is then calculated by the combination of maximum softmax value and entropy as described previously, both with temperature scaling. 
\begin{eqnarray}
 \hat{x_t} &=& x_t - \epsilon*\text{sign}(\frac{\partial L(F_i(x_t;T))}{\partial x_t})
\label{preprocessing}
\end{eqnarray}

The complete algorithm for OOD detection on a test image is presented in Algorithm~\ref{algo3}.

\section{Experimental Results}
\label{sec:experiments}
In this section, we describe our experimental results. The details such as in-distribution and OOD datasets, the neural network architectures and evaluation metrics are described in detail. The ablation studies on various hyper-parameters of the algorithms are described and conclusions are drawn. Finally, our method is compared against the current state-of-the-art ODIN~\cite{liang2018odin} and is shown to significantly outperform it.

\begin{table}[t]
\caption{Test error rates on CIFAR-10 and CIFAR-100}
\label{tab:test-err}
\centering
\begin{tabular}{l | l |l }
  \toprule
  {\bf Architecture} & {\bf CIFAR-10} & {\bf CIFAR-100}\\ 
  \midrule
  \hline
 DenseNet-BC & 5.0 & 19.9\\ 
 \hline
 WRN-28-10  & 5.0 & 20.4\\ 
    \bottomrule
\end{tabular}
\end{table}

\subsection{Experimental Setup}

We use CIFAR-10 (contains 10 classes) and CIFAR-100 (contains 100 classes)~\cite{krizhevsky2009cifar} datasets as in-distribution datasets to train deep neural networks for image classification. They both consist of 50,000 images for training, and 10,000 images for testing. The dimensions of an image in both the datasets is $32 \times 32$. The classes of both CIFAR-10 and CIFAR-100 are randomly divided into five parts. As described in Section~\ref{sec:leave-out}, each part is assigned as OOD to a unique network which is then trained. For each network, the other parts act as in-distribution samples. 

Following the benchmarks given in \cite{liang2018odin}, the following OOD datasets are used in our experiments. The datasets are described in ODIN\cite{liang2018odin} and provided as a part of their code release; here we are simply restating the description for comprehensiveness.
\begin{itemize}
\item {\bf TinyImageNet\cite{tinyimagenet} \newtext{(TIN)}} is a subset of ImageNet dataset\cite{liang2018odin}. Tiny  ImageNet contains 200 classes which is drawn from original 1,000 classes of ImageNet. In total, there are 10,000 images in the Tiny ImageNet. By randomly cropping and downsampling each image  to $32 \times 32$, two datasets \newtext{{\it TinyImageNetcrop} ({\it TINc}) and {\it TinyImageNetresize} ({\it TINr})} are constructed.

\item {\bf LSUN} is the Large Scale UNderstanding dataset (LSUN)\cite{yu2015lsun} created by Princeton, using deep learning classifiers with humans in the loop. It contains 10,000 images of 10 scene categories. By randomly cropping and downsampling each image to size $32 \times 32$, two datasets \newtext{{\it LSUNc} and {\it LSUNr}} are constructed.

\item {\bf iSUN\cite{xu2015isun}} is collected by gaze tracking from Amazon Mechanical Turk using a webcam. It contains 8925 scene images. Similar to the above dataset as other datasets, images are down-sampled to size $32 \times 32$.

\item {\bf Uniform Noise \newtext{(UNFM)}} is synthetic dataset consists of 10,000 noise images. The RGB value of each pixel in an image is drawn from uniform distribution in the range $[0,1]$.

\item {\bf Gaussian Noise \newtext{(GSSN)}} is synthetic dataset consists of 10,000 noise images. The RGB value of each pixel is drawn from independent and identically distributed Gaussian with mean 0.5 and unit variance and each pixel value is clipped to the range $[0,1]$.

\end{itemize}

{\bf Neural network architecture} Following ODIN\cite{liang2018odin}, two state-of-the-art neural network architectures, {\it DenseNet}~\cite{huang2016densenet} and {\it Wide ResNet}(WRN)~\cite{zagoruyko2016wrn}, are adopted to evaluate our method. For DenseNet, we use the DenseNet-BC setup as in \cite{huang2016densenet}, with depth $L=100$, growth rate $k=12$ and dropout rate 0. For Wide ResNet, we use WRN-28-10 setup, with depth 28, width 10 and dropout rate of 0.3. We train both DenseNet-BC and Wide ResNet on CIFAR-10 and CIFAR-100 for 100 epochs with batch size 100, momentum 0.9, weight decay 0.0005, and margin 0.4. The initial training rate is 0.1 and it is linearly dropped to 0.0001 over the whole training process. During training, we augment our training data with random flip and random cropping. \newtext{We use the smallest OOD dataset, iSUN, as validation data for hyper-parameter search}. We test \newtext{the rest four} out-of-distribution datasets \newtext{except iSUN} on our trained network. During testing, we use batch size 100. Similar to ODIN\cite{liang2018odin}, input preprocessing with $\epsilon=0.002$ is used.

Table~\ref{tab:test-err} shows the test error rates when our method is trained and tested on CIFAR-10 and CIFAR-100 respectively using the algorithms~\ref{algo2} and~\ref{algo3}. For CIFAR-10, the vanilla DenseNet-BC~\cite{huang2016densenet} and the proposed method gives error rates of $4.51\%$ and $5.0\%$ respectively. For CIFAR-100, the error rates are $22.27\%$ and $19.9\%$ respectively. On both these datasets, the difference in error rates is marginal. 
For WRN~\cite{zagoruyko2016wrn} with depth 40, $k=10$, the test error rate on CIFAR-10 for the vanilla network is $4.17\%$ and for the proposed network is $5.0\%$. For CIFAR-100, the error rates are $20.5\%$ and $20.4\%$ for the vanilla network and the proposed network respectively. The small difference in the performance on CIFAR-10 can be explained by the fact that the our method did not use the ZCA whitening preprocessing while the vanilla network did.

\subsection{Evaluation Metrics}
To measure the effectiveness of our method to distinguish between in-distribution and out-of-distribution samples, we adopt five different metrics, same as what was used in ODIN\cite{liang2018odin} paper. We restate these metrics below for comprehensiveness. In the rest of manuscript, TP, TN, FP, FN are used to denote true positives, true negatives, false positives and false negatives respectively.



{\bf FPR at 95\% TPR} measures the probability that an out-of-distribution sample is misclassified as in-distribution when the true positive rate (TPR) is 95\%. In this metric, TPR is computed by $TP/(TP+TN)$, and FPR is computed by $FP/(FP+TN)$.

{\bf Detection Error} measures the minimum misclassification probability over all possible score thresholds, as defined in ODIN\cite{liang2018odin}. To have a fair comparison with ODIN, the same number of positive and negative samples are used during testing.

{\bf AUROC} is the Area Under the Receiver Operating Characteristic curve. In a ROC curve, the TPR is plotted as a function of FPR for different threshold settings. AUROC equals to the probability that a classifier will rank a randomly chosen positive sample higher than a randomly chosen negative one. AUROC score of 100\% means perfect separation between positive and negative samples. 

{\bf AUPR-In} measures the Area Under the Precision-Recall curve. In a PR curve, the {\it precision}$=TP/(TP+FP)$, is plotted as a function of {\it recall}$=TP/(TP+FN)$, for different threshold settings. Since precision is directly influenced by class imbalance (due to FP), PR curves can highlight performance differences that are lost in ROC curves for imbalanced datasets\cite{Goadrich2006aupr}. AUPR score of 100\% means perfect distinguish between positive and negative samples. For AUPR-In metric, in-distribution images are specified as positive.

{\bf AUPR-Out} is similar to the metric AUPR-In. The difference lies in that for AUPR-Out metric, out-of-distribution images are specified as positive.


{\bf CLS Acc} is the classification accuracy for ID samples. \\

\subsection{Ablation Studies}

\begin{table}[t]
\fontsize{6}{9}\selectfont
\begin{center}
\begin{tabularx}{\textwidth}{p{0.7cm} p{2.2cm} s s s s s s}
\toprule 
\begin{tabular}{@{}c@{}}\bf{Ablation} \\ \bf{Studies} \end{tabular}
& \centering{\bf Parameters}
& \begin{tabular}{@{}c@{}} \bf{FPR at} \\ \bf{95\% TPR} \\ \bf{$\downarrow$} \end{tabular} 
& \begin{tabular}{@{}c@{}} \bf{Detection} \\ \bf{Error} \\ \bf{$\downarrow$} \end{tabular} 
& \begin{tabular}{@{}c@{}} \bf{AUROC} \\  \\ \bf{$\uparrow$} \end{tabular}
& \begin{tabular}{@{}c@{}} \bf{AUPR} \\ \bf{In} \\ $\uparrow$ \end{tabular}
&  \begin{tabular}{@{}c@{}} \bf{AUPR} \\ \bf{Out} \\ $\uparrow$ \end{tabular}
&  \begin{tabular}{@{}c@{}} \bf{CLS} \\ \bf{Acc} \\ $\uparrow$ \end{tabular}\\ 
 \hline
\parbox[t]{1mm}{\multirow{4}{*}{\rotatebox[origin=c]{90}{\begin{tabular}{@{}c@{}} \bf{ Number}\\ \bf{of} \\ \bf{Splits}   \end{tabular}}}}
& \centering{3} &32.37 &13.94 &93.50 &94.39 &92.22 &76.41 \\
& \centering{5}  &{\bf22.95} &{\bf10.79} &{\bf95.69} &{\bf96.55} &94.3 &80.01 \\
& \centering{10} &28.71 &12.53 &94.48 &95.37 &93.26 &81.94 \\
& \centering{20} &23.85 &10.95 &95.49 &96.24 &{\bf94.36} &{\bf82.33}\\
 \hline
\parbox[t]{1mm}{\multirow{3}{*}{\rotatebox[origin=c]{90}{\begin{tabular}{@{}c@{}} \bf{ Type}\\ \bf{of} \\ \bf{splits}  \end{tabular}}}}
&  &  &  &  &  & &  \\
& \centering{Random}  &{\bf22.95} &{\bf10.79} &{\bf95.69} &{\bf96.55} &{\bf94.3} &{\bf80.01} \\
& \centering{Manual} &40.16 &16.26 &91.57 &92.90 &89.57 &79.79\\
 \hline
 \parbox[t]{1mm}{\multirow{6}{*}{\rotatebox[origin=c]{90}{\begin{tabular}{@{}c@{}}  \bf{Epsilon}  \end{tabular}}}}
& \centering{0.000000}&53.51 &16.37 &90.75 &93.16 &86.71 &{\bf80.32} \\
& \centering{0.000313}&41.62 &14.37 &92.8 &94.52 &90.16 &80.22 \\
& \centering{0.000625} &34.64 &12.83 &94.09 &95.4 &92.19 &80.17 \\
& \centering{0.001250}&25.74 &11.19 &95.38 &96.31 &94.04 &80.08 \\
& \centering{0.002000}&{\bf22.95} &{\bf10.79} &{\bf95.69} &{\bf96.55} &{\bf94.3} &80.01\\
& \centering{0.003000}&29.07 &11.79 &94.73 &95.9 &92.43 &79.97 \\
 \hline
 \parbox[t]{1mm}{\multirow{5}{*}{\rotatebox[origin=c]{90}{\begin{tabular}{@{}c@{}}  \bf{Temp-} \\ \bf{rature} \end{tabular}}}}
& \centering{1}&38.57 &17.32 &91.44 &92.7 &90.12 &{\bf80.01} \\
& \centering{10}&27.84 &11.93 &94.86 &95.81 &93.39 &{\bf80.01} \\
& \centering{100}&24.44 &10.86 &95.6 &96.5 &94.17 &{\bf80.01} \\
& \centering{1000}&22.95 &{\bf10.79} &{\bf95.69} &{\bf96.55} &{\bf94.3} &{\bf80.01} \\
& \centering{5000}&{\bf22.7} &10.81 &95.66 &96.53 &94.28 &{\bf80.01} \\
 \hline
 \parbox[t]{1mm}{\multirow{4}{*}{\rotatebox[origin=c]{90}{\begin{tabular}{@{}c@{}} \bf{Loss} \\ \bf{Function} \end{tabular}}}}
&  &  & &  &  & & \\
& \centering{SFX} &84.09 &36.55 &68.96 &72.38 &63.77 &54.18 \\
& \centering{SFX+MaxEntropyDiff} &50.70 &19.65 &88.26 &89.71 &86.18 &72.99\\
& \centering{SFX+MarginEntropy} &{\bf22.95} &{\bf10.79} &{\bf95.69} &{\bf96.55} &{\bf94.3} &{\bf80.01}\\
 \hline
 \parbox[t]{1mm}{\multirow{6}{*}{\rotatebox[origin=c]{90}{\begin{tabular}{@{}c@{}} \bf{OOD} \\ \bf{Detection} \\ \bf{Score} \end{tabular}}}}
& \centering{SFX} &50.52 &19.91 &88.69 &90.91 &86.19 &{\bf80.01}\\
& \centering{Entropy} &36.23 &16.48 &91.92 &93.03 &90.74 &{\bf80.01}\\
& \centering{SFX+Entropy} &38.57 &17.32 &91.44 &92.7 &90.12 &{\bf80.01}\\
& \centering{SFX@Temp} &{\bf22.71} &10.83 &95.65 &96.52 &94.26 &{\bf80.01}\\
& \centering{Entropy@Temp}&37.0 &14.05 &93.33 &94.76 &91.04 &{\bf80.01}\\
& \centering{(SFX+Entropy)@Temp} &22.95 &{\bf10.79} &{\bf95.69} &{\bf96.55} &{\bf94.3} &{\bf80.01}\\
\bottomrule
\end{tabularx}
\caption{Ablation Studies on CIFAR-100 as in-distribution data and iSUN as out-of-distribution data on DenseNet-100 network. All values are percentages. $\uparrow$ indicates larger value is better, and $\downarrow$ indicates lower value is better.}
\label{tab:ablation}
\end{center}
\end{table}

In this section, we perform ablation studies to study the effects of various hyper parameters used in our model. We perform the ablation studies on DenseNet-BC~\cite{huang2016densenet} network with CIFAR-100~\cite{krizhevsky2009cifar} as in-distribution while training and iSUN~\cite{xu2015isun} as the OOD \newtext{validation} data while testing. By default, we use $5$ random splits for CIFAR-100,  $\epsilon=0.002$,  \emph{ SFX+MarginEntropy} loss to train network, \deltatext{accuracy bound $\delta=2 \%$ to save the models}, and use  \emph{(Softmax + Entropy)@Temperature} with $T=1000$ to detect out-of-distribution samples. Results are given in Table~\ref{tab:ablation}.

{(1) \bf Number of splits: } This analysis characterizes the sensitivity of our algorithm to the number of splits of the training classes which is same as the number of classifiers in the ensemble. As the number of splits increase, the number of times a particular training class being in-distribution for the leave-out classifiers increases too. This enables the ensemble to discriminate an in-distribution sample from the OOD sample. But it also increases the computational cost. For CIFAR-100, we studied 3, 5, 10 and 20 splits. Our results show that while 5 splits gave the best result, 3 splits also provides a good trade-off between accuracy and computational cost. We choose the number of splits as 5 as default value.

\vspace{3pt}
{\bf (2) Type of splits:} This study characterizes the way in which the classes are split into mutually exclusive sets. We experiment with splitting the classes manually using prior knowledge and splitting randomly. For the manual split, the class labels are first clustered into semantically consistent groups and classes from each group are then distributed across the splits. The results show that the OOD detection rates for random selections are better than the manual selection.\newtext{This ensures that we can achieve good OOD detection rates even by random selection of classes when the number of classes is huge.}



\vspace{3pt}

{\bf (3) Different $\epsilon$ for input preprocessing: } For input preprocessing, we sweep over $\epsilon \in [0,0.000313,0.000625,0.00125,0.002,0.003]$. Our results show that as $\epsilon$ increases from 0, the performance of out-of-distribution detector increases, and it reaches the best performance at $0.002$. The further increase of $\epsilon$ does not help performance.
\vspace{3pt}

\newtext{{\bf (4) Different $T$ for temperature scaling: } For temperature scaling, we sweep over $T \in [1, 10, 100, 1000, 5000]$. Our results show that for DenseNet-BC with CIFAR-100, as $T$ increases from 1 to 1000, the performance of out-of-distribution detector increases. Beyond $T=1000$, the performance does not change much.} 
\vspace{3pt}

{\bf (5) Loss function variants:}
We study the effects of training \newtext{our method} with different types of losses. \newtext{The training regime follows the strategy given in section~\ref{sec:leave-out}, where the training data $X$ is split into $K$ partitions $X_i,i\in \{1,...,K\}$. A total of $K$ classifiers are trained where classifier $F_i$ uses the partition $X_i$ as OOD data and $X-X_i$ as in-distribution data.}
\begin{itemize}
\item \emph{ SFX}: We assign an additional label to all the OOD samples and train classification network using the cross entropy loss. 
\item { \emph{ SFX+MaxEntropyDiff}: Along with the cross entropy loss, we maximize the difference between the entropy of in- and out-of-distribution samples across all in-distribution classes.}
\item {\emph{ SFX+MarginEntropy}: Along with the cross entropy, we maximize the difference between the entropy of in- and out-of-distribution samples across all in-distribution classes, but is bounded by a margin as given in Equation~\ref{loss}. }
\end{itemize}
Our results show that the proposed \emph{SFX+MarginEntropy} loss works dramatically better than all other types of losses for detecting out-of-distribution samples \newtext{as well as for accurate classification}. The results demonstrate that the proposed novel loss function \emph {SFX+MarginEntropy} (equation~\ref{loss}) is the major factor in significant improvements over the current state-of-the-art ODIN~\cite{liang2018odin}.

\vspace{3pt}
{\bf (6) Out-of-distribution detector:}
We study different OOD scoring methods to discriminate out-of-distribution samples from in-distribution samples.
\begin{itemize}
\item { \emph{ Softmax score}: Given an input image, the score is given by the average of maximum softmax outputs over all the classifiers in the ensemble.}
\item { \emph{ Entropy score}: Given an input image, the score is given by  the average of entropy of softmax vector over all the classifiers in the ensemble.}
\item { \emph{ Softmax + Entropy}: Given an input image, both the above scores are added.}
\item { \emph{ Softmax@Temperature}: Given an input image, the above described {\emph Softmax score} is computed on temperature scaled ($T=1000$) softmax vectors.}
\item { \emph{Entropy@Temperature}: Given an input image, the above described {\emph Entropy score} is computed on temperature scaled ($T=1000$) softmax vectors.}
\item { \emph{(Softmax + Entropy)@Temperature}: Given an input image, the above described {\emph Entropy@Temperature} and {\emph Softmax@Temperature} are computed ($T=1000$) on softmax vectors and then added.}
\end{itemize}
Among the above OOD scoring methods, \newtext{the \emph{(Softmax + Entropy)@Temperature} ($T=1000$) achieved the best performance. \emph{ Softmax@Temperature} ($T=1000$) achieved the second best performance.}


\subsection{Results and Analysis}

\begin{table}[t]
\fontsize{7}{9}\selectfont
\begin{center}
\begin{tabularx}{\textwidth}{p{0.8cm}  p{1.0cm} s s s s s}
\toprule 
& \begin{tabular}{@{}c@{}}\bf{OOD} \\ \bf{Dataset} \end{tabular}  
& \begin{tabular}{@{}c@{}} \bf{FPR at} \\ \bf{95\% TPR} \\ \bf{$\downarrow$} \end{tabular} 
& \begin{tabular}{@{}c@{}} \bf{Detection} \\ \bf{Error} \\ \bf{$\downarrow$} \end{tabular} 
& \begin{tabular}{@{}c@{}} \bf{AUROC} \\  \\ \bf{$\uparrow$} \end{tabular}
& \begin{tabular}{@{}c@{}} \bf{AUPR} \\ \bf{In} \\ $\uparrow$ \end{tabular}
&  \begin{tabular}{@{}c@{}} \bf{AUPR} \\ \bf{Out} \\ $\uparrow$ \end{tabular}\\ 
 \hline
& & \multicolumn{5}{c}{each cell in {\bf ODIN\cite{liang2018odin}/Our Method} format} \\
\cline{3-7}
\parbox[t]{1mm}{\multirow{6}{*}{\rotatebox[origin=c]{90}{\begin{tabular}{@{}c@{}} \bf{DenseNet-BC} \\ \bf{CIFAR-10} \end{tabular}}}}
& TINc & 4.30/{\bf 1.23} & 4.70/{\bf 2.63}& 99.10/{\bf 99.65} & 99.10/{\bf 99.68} & 99.10/{\bf 99.64} \\
& TINr & 7.50/{\bf 2.93} & 6.10/{\bf 3.84}& 98.50/{\bf 99.34} & 98.60/{\bf 99.37} & 98.50/{\bf  99.32} \\
 & LSUNc & 8.70/{\bf 3.42} & 6.00/{\bf  4.12}& 98.20/{\bf 99.25} & 98.50/{\bf 99.29} & 97.80/{\bf 99.24} \\
& LSUNr & 3.80/{\bf 0.77} & 4.40/{\bf 2.1}& 99.20/{\bf 99.75} & 99.30/{\bf 99.77} & 99.20/{\bf 99.73} \\
& UNFM & {\bf 0.00}/2.61 & {\bf0.20}/3.6 & {\bf 100}/98.55 & {\bf 100}/98.94 & {\bf 100}/97.52 \\
& GSSN & {\bf0.00}/{\bf 0.00} & {\bf0.50}/0.2& {\bf99.90}/99.84 & {\bf 100}/99.89 & {\bf99.90}/99.6\\
 \hline
 \parbox[t]{1mm}{\multirow{6}{*}{\rotatebox[origin=c]{90}{\begin{tabular}{@{}c@{}} \bf{DenseNet-BC} \\ \bf{CIFAR-100} \end{tabular}}}}
& TINc & 17.30/{\bf 8.29} & 8.80/{\bf 6.27} & 97.10/{\bf98.43} & 97.40/{\bf98.58} & 96.80/{\bf98.3} \\
& TINr & 44.30/{\bf20.52} & 17.50/{\bf9.98} & 90.70/{\bf96.27} & 91.40/{\bf96.66} & 90.10/{\bf95.82} \\
& LSUNc & 17.60/{\bf14.69} & 9.40/{\bf8.46} & 96.80/{\bf97.37} & 97.10/{\bf97.62} & 96.50/{\bf97.18} \\
 & LSUNr & 44.00/{\bf16.23} & 16.80/{\bf8.77} & 91.50/{\bf97.03} & 92.40/{\bf97.37} & 90.60/{\bf96.6} \\
& UNFM & {\bf0.50}/79.73 & {\bf2.50}/9.46 & {\bf99.50}/92.0 & {\bf99.60}/94.77 & {\bf99.00}/83.81 \\
& GSSN & {\bf0.20}/38.52 & {\bf1.90}/8.21 & {\bf99.60}/94.89 & {\bf99.70}/96.36 & {\bf99.10}/90.01 \\
\hline
 \parbox[t]{1mm}{\multirow{6}{*}{\rotatebox[origin=c]{90}{\begin{tabular}{@{}c@{}} \bf{WRN-28-10} \\ \bf{CIFAR-10} \end{tabular}}}}
& TINc & 23.40/{\bf0.82 } & 11.60/{\bf2.24 }& 94.20/{\bf99.75 } & 92.80/{\bf99.77 } & 94.70/{\bf99.75 } \\
& TINr & 25.50/{\bf2.94 } & 13.40/{\bf3.83 }& 92.10/{\bf99.36} & 89.00/{\bf99.4} & 93.60/{\bf99.36 } \\
 & LSUNc & 21.80/{\bf1.93 } & 9.80/{\bf3.24 }& 95.90/{\bf99.55 } & 95.80/{\bf99.57 } & 95.50/{\bf99.55 } \\
& LSUNr & 17.60/{\bf0.88 } & 9.70/{\bf2.52 }& 95.40/{\bf99.7 } & 93.80/{\bf99.72 } & 96.10/{\bf99.68 } \\
& UNFM & {\bf0.00}/16.39 & {\bf0.20}/5.39 & {\bf100}/96.77 & {\bf 100}/97.78 & {\bf100}/94.18 \\
& GSSN & {\bf 0.00}/{\bf 0.00} & {\bf 0.10}/1.03 & {\bf 100}/99.58 & {\bf 100}/99.71 & {\bf 100}/99.2 \\
 \hline
 \parbox[t]{1mm}{\multirow{6}{*}{\rotatebox[origin=c]{90}{\begin{tabular}{@{}c@{}} \bf{WRN-28-10} \\ \bf{CIFAR-100} \end{tabular}}}}
& TINc & 43.90/{\bf 9.17} & 17.20/{\bf 6.67} & 90.80/{\bf 98.22} & 91.40/{\bf 98.39} & 90.00/{\bf 98.07} \\
& TINr & 55.90/{\bf 24.53} & 23.30/{\bf 11.64} & 84.00/{\bf 95.18} & 82.80/{\bf 95.5} & 84.40/{\bf 94.78} \\
& LSUNc & 39.60/{\bf14.22} & 15.60/{\bf8.2} & 92.00/{\bf97.38} & 92.40/{\bf97.62} & 91.60/{\bf97.16} \\
& LSUNr & 56.50/{\bf16.53} & 21.70/{\bf9.14} & 86.00/{\bf 96.77} & 86.20/{\bf 97.03} & 84.90/{\bf 96.41} \\
& UNFM & {\bf0.10}/99.9 & {\bf2.20}/14.86 & {\bf99.10}/83.44 & {\bf99.40}/89.43 & {\bf97.50}/71.2 \\
& GSSN & {\bf1.00}/98.26 & {\bf2.90}/16.88 & {\bf98.50}/93.04 & {\bf99.10}/88.64 & {\bf95.90}/71.62 \\
\hline
\bottomrule
\end{tabularx}
\caption{Distinguishing in- and out-of-distribution test set data for the image classification. All values are percentages. $\uparrow$ indicates larger value is better, and $\downarrow$ indicates lower value is better. Each value cell is in "ODIN\cite{liang2018odin}/Our Method" format.}
\label{tab:main}
\end{center}
\end{table}

\begin{table}[t]
\fontsize{7}{9}\selectfont
\begin{center}
\begin{tabularx}{\textwidth}{p{1.2cm} s s s s }
\toprule 
\begin{tabular}{@{}c@{}}\bf{OOD} \\ \bf{Dataset} \end{tabular}  
& \begin{tabular}{@{}c@{}} \bf{DenseNet-BC} \\ \bf{CIFAR-10} \end{tabular} 
& \begin{tabular}{@{}c@{}} \bf{DenseNet-BC} \\ \bf{CIFAR-100} \end{tabular} 
& \begin{tabular}{@{}c@{}} \bf{WRN-28-10} \\ \bf{CIFAR-10}   \end{tabular}
& \begin{tabular}{@{}c@{}} \bf{WRN-28-10} \\ \bf{CIFAR-100} \end{tabular}\\ 
 \hline
TINc &$1.49\pm0.23$ &$10.26\pm1.33$ &$1.3\pm0.33$ &$10.35\pm2.21$ \\
TINr &$3.95\pm0.75$ &$26.58\pm4.16$ &$4.56\pm1.29$ &$29.84\pm5.12$ \\
LSUNc &$4.54\pm1.42$ &$16.95\pm1.27$ &$3.81\pm1.22$ &$15.51\pm1.4$ \\
LSUNr &$1.3\pm0.6$ &$20.22\pm2.79$ & $1.53\pm0.41$&$22.51\pm6.08$ \\
UNFM &$14.37\pm31.84$ &$38.79\pm19.41$ &$0.75\pm1.24$ &$47.67\pm47.19$ \\
GSSN &$27.09\pm40.02$ &$82.24\pm12.81$ & $31.47\pm33.95$&$67.48\pm44.33$ \\
 
\bottomrule
\end{tabularx}
\caption{Mean and standard deviation of FPR at 95\% TPR}
\label{tab:main-std}
\end{center}
\end{table}

\begin{figure*}[t]
\centering
\subfloat[\small Cross entropy loss.\label{fig:cross-entropy}]{%
  \centering
  \includegraphics[width=0.45\columnwidth,height=3.5cm]{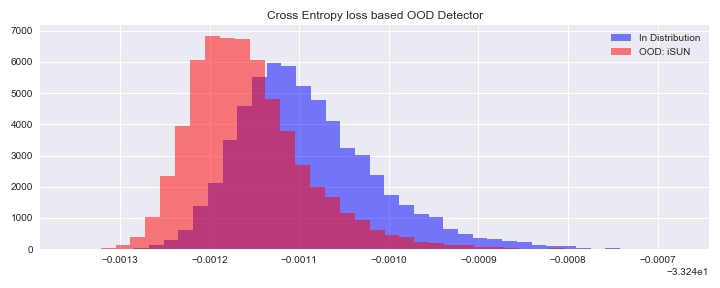}
  }
  \subfloat[\small Margin entropy loss.\label{fig:margin-entropy}]{%
  \centering
  \includegraphics[width=0.45\columnwidth,height=3.5cm]{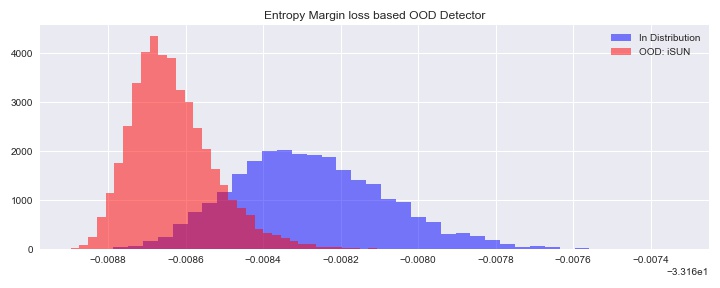}
  }
  \caption{\small Histogram of ID and OOD detection scores of the proposed method and ODIN~\cite{liang2018odin}}
  \label{fig:loss-diff}
\end{figure*}

Table~\ref{tab:main} shows the comparison between our results and ODIN~\cite{liang2018odin} on various benchmarks. 
The results are reported on all neural network, in-dataset and OOD dataset combinations. Our hyperparameters are tuned using iSUN dataset. From the Table~\ref{tab:main}, it is very clear that our approach significantly outperforms ODIN~\cite{liang2018odin} across all neural network architectures on almost all of the dataset pairs. The combination of novel loss function, OOD scoring method, and the ensemble of models has enabled our method to significantly improve the performance of OOD detection on more challenging datasets, such as LSUN (resized), iSUN and ImageNet(resized), where the images contain full objects as opposed to the cropped parts of objects. The proposed method is slightly worse on the uniform and some of Gaussian distribution results. 
Moreover, our method achieves significant gains on both CIFAR-10 and CIFAR-100 with the same number of splits which is $5$, even though the number of classes have increased by a factor of ten from CIFAR-10 to CIFAR-100. Thus the number of splits need not be scaled linearly with the number of classes, making our method practical. We implicitly outperform Hendrycks and Gimpel~\cite{hendrycks2017baseline} and Lee \emph{et.al,}~\cite{lee2018training} as ODIN outperforms both these works and our method outperform ODIN on all but two benchmarks. 

All three components in our method, namely novel loss function, the ensemble of leave-out classifiers and improved OOD detection metric contributed to the improvement in performance over state-of-the-art ODIN (refer to table~\ref{tab:ablation}). 
The contribution of these components can be seen in Table~\ref{tab:ablation} in the rows marked as ``Loss function'', ``Number of splits'' and ``OOD detection scores''.


\newtext{Our algorithm has stochasticity in the form of random splits of the classes. Given 100 classes in CIFAR-100, there are many ways to split 100 classes into 5 partitions. Table~\ref{tab:main-std} gives the mean and standard deviation across five random ways to partition data when we use 5 number of splits for training. We note that even our worst case results outperform ODIN~\cite{liang2018odin} on more challenging datasets.}

\begin{figure}[t]
\centering
\subfloat[\small ImageNet.\label{fig:imagenet}]{%
  \centering
  \includegraphics[width=0.5\columnwidth,height=3cm]{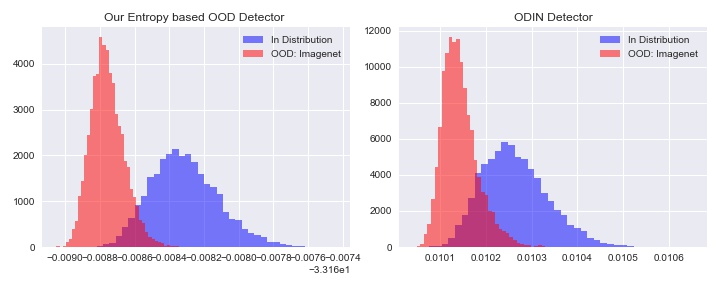}
  }
\subfloat[\small resized ImageNet.\label{fig:imagenet_resize}]{%
  \centering
  \includegraphics[width=0.5\columnwidth,height=3cm]{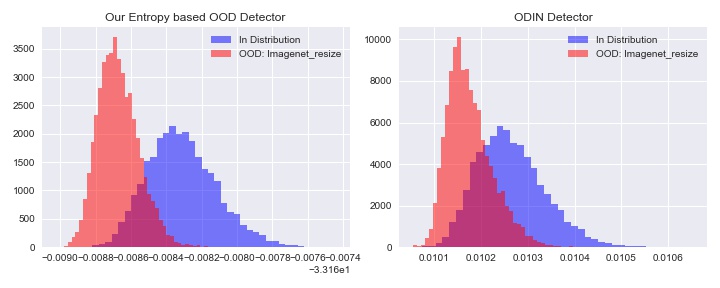}
  } \\
  \subfloat[\small LSUN.\label{fig:lsun}]{%
  \centering
  \includegraphics[width=0.5\columnwidth,height=3cm]{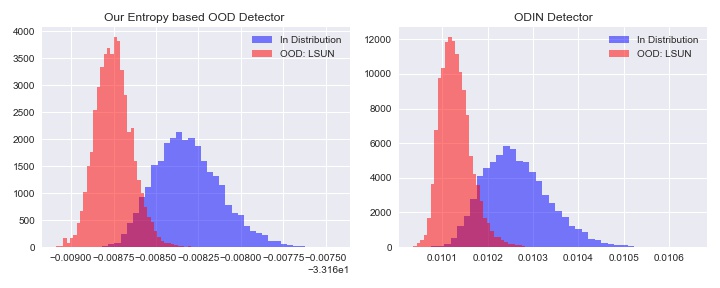}
  }
\subfloat[\small resized LSUN.\label{fig:lsun_resize}]{%
  \centering
  \includegraphics[width=0.5\columnwidth,height=3cm]{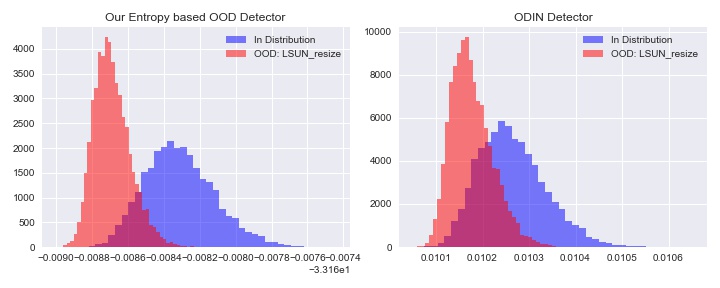}
  }
  \caption{\small Histogram of ID and OOD detection scores with proposed OOD detector v.s. ODIN OOD detector}
  \label{fig:ood-detect}
\end{figure}

\newtext{Figure~\ref{fig:loss-diff} compares the histogram of OOD detection scores on ID and OOD samples when different loss functions are used for training. Figure~\ref{fig:loss-diff}(a) is trained with only cross entropy loss, while Figure~\ref{fig:loss-diff}(b) is trained with proposed margin entropy loss and cross entropy, the proposed OOD detector is used in both bases. As shown in Figure~\ref{fig:loss-diff}, the proposed margin entropy loss helps to better separate ID and OOD distributions than using cross entropy loss alone.}
Figure~\ref{fig:ood-detect} presents the histogram plot of OOD detection scores on ID and OOD samples for both our method and ODIN~\cite{liang2018odin}. As shown in Figure~\ref{fig:ood-detect}, the proposed method has less overlap between OOD samples and ID samples compared to ODIN~\cite{liang2018odin} and thus separates ID and OOD distributions better.

\section{Conclusion and Future Work}
\label{sec:conclusion}
As deep learning is widely adopted in many commercially important applications, it is very important that anomaly detection algorithms are developed for these algorithms. In this work, we have proposed an anomaly detection algorithm for deep neural networks which is an ensemble of leave-out-classifiers. These classifiers are learned by maximizing the margin-loss between the entropy of OOD samples and in-distribution samples. A random subset of training data serves as OOD data while the rest of the data serves as in-distribution. We show our algorithm  significantly outperforms the current state-of-art methods~\cite{hendrycks2017baseline}, ~\cite{lee2018training} and~\cite{liang2018odin} across almost all the benchmarks. Our method contains three important components, novel loss function, the ensemble of leave-out classifiers, and novel out-of-distribution detector. Each of this component improves OOD detection performance. Each of them can be applied independently on top of other methods.

We also note that this method opens up several directions of research to pursue. First, the proposed method of the ensemble of neural networks requires large memory and computational resources. This can potentially be alleviated by all the networks sharing most of the parameters and branch away individually. \newtext{Also, the number of splits can be used to trade off between detection performance and computational overhead. Notice that based on ablation study (Table~\ref{tab:ablation}) and detailed 3 splits results in supplementary document, even 3 splits outperform ODIN~\cite{liang2018odin}. For use cases where reducing computational time is critical, we recommend to use 3 splits. Please see supplementary material for detailed results on 3 splits.} Our current work requires an OOD dataset \newtext{for} hyper-parameter search. This problem can potentially be solved by investigating other surrogate functions for entropy which are better behaved with the epochs. 

\bibliographystyle{splncs04}
\bibliography{2698}

\end{document}